\documentclass{article}

\usepackage[english]{babel}

\usepackage[letterpaper,top=2cm,bottom=2cm,left=3cm,right=3cm,marginparwidth=1.75cm]{geometry}

\usepackage{amsmath}
\usepackage{graphicx}
\usepackage[colorlinks=true, allcolors=blue]{hyperref}
\usepackage{filecontents}
\usepackage{float}

\title{A Mixture of Experts Gating Network for Enhanced Surrogate Modeling in External Aerodynamics}

\author{Mohammad Amin Nabian, Sanjay Choudhry \\
NVIDIA}

\begin{document}
\maketitle

\begin{abstract}
The computational cost associated with high-fidelity Computational Fluid Dynamics (CFD) simulations remains a significant bottleneck in the automotive design and optimization cycle. While machine learning (ML) based surrogate models have emerged as a promising alternative to accelerate aerodynamic predictions, the field is characterized by a diverse and rapidly evolving landscape of specialized neural network architectures, with no single model demonstrating universal superiority. This paper introduces a novel meta-learning framework that leverages this architectural diversity as a strength. We propose a Mixture of Experts (MoE) model that employs a dedicated gating network to dynamically and optimally combine the predictions from three heterogeneous, state-of-the-art surrogate models: DoMINO, a decomposable multi-scale neural operator; X-MeshGraphNet, a scalable multi-scale graph neural network; and FigConvNet, a factorized implicit global convolution network. The gating network learns a spatially-variant weighting strategy, assigning credibility to each expert based on its localized performance in predicting surface pressure and wall shear stress fields. To prevent model collapse and encourage balanced expert contributions, we integrate an entropy regularization term into the training loss function. The entire system is trained and validated on the DrivAerML dataset, a large-scale, public benchmark of high-fidelity CFD simulations for automotive aerodynamics. Quantitative results demonstrate that the MoE model achieves a significant reduction in L-2 prediction error, outperforming not only the ensemble average but also the most accurate individual expert model across all evaluated physical quantities. This work establishes the MoE framework as a powerful and effective strategy for creating more robust and accurate composite surrogate models by synergistically combining the complementary strengths of specialized architectures.

\begin{figure}[H]
\centering
\includegraphics[width=\textwidth]{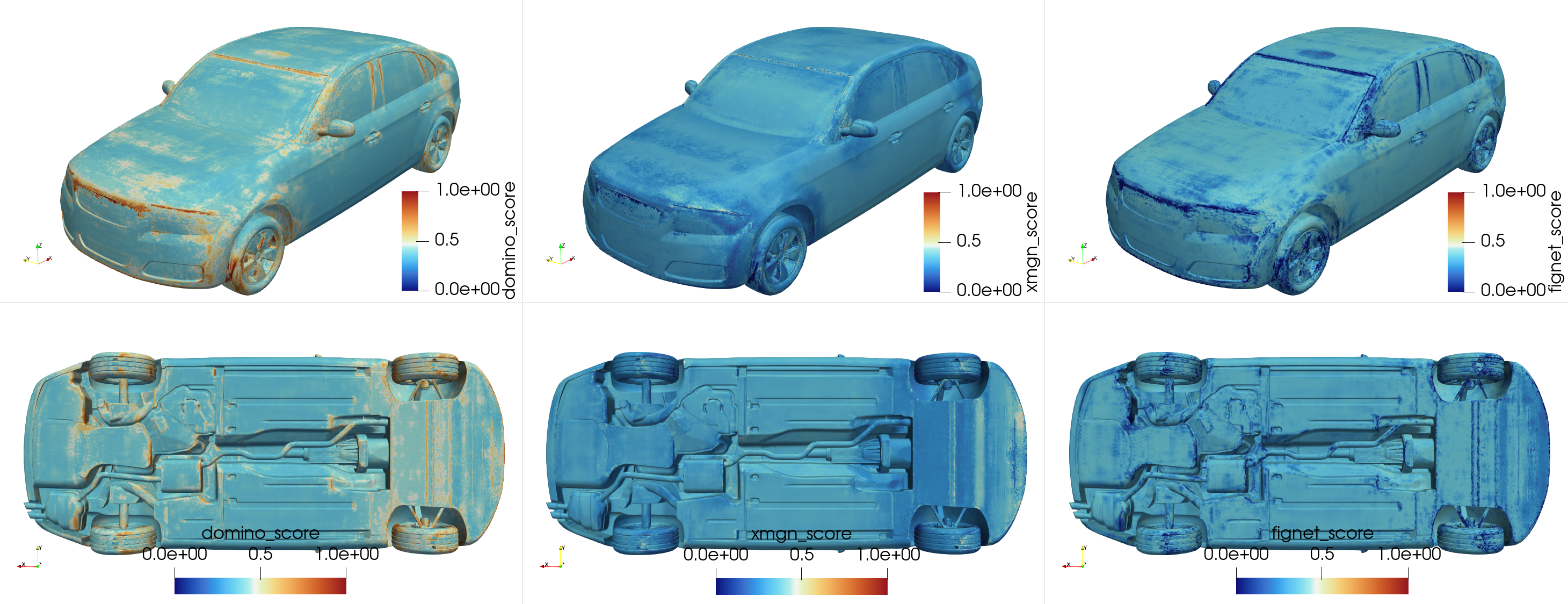} 
\caption{Pointwise MoE gating weights for pressure prediction on a sample from the DrivAerML dataset. The visualization reveals a physically meaningful, spatially-variant weighting strategy learned by the gating network.}
\label{fig:weights}
\end{figure}

\end{abstract}

\section{Introduction}
\subsection{The Computational Grand Challenge in Automotive Aerodynamics}

The design of modern road vehicles is a complex, multi-objective optimization problem where external aerodynamics plays a pivotal role \cite{ashton2024drivaerml}. Aerodynamic forces, such as drag and lift, directly influence critical performance metrics including fuel efficiency for internal combustion engines, range for electric vehicles, and high-speed stability \cite{choy2025factorized}. Furthermore, the turbulent airflow around a vehicle is a primary source of aeroacoustic noise, a key factor in passenger comfort \cite{ashton2024drivaerml}. For decades, Computational Fluid Dynamics (CFD) has been an indispensable tool in the automotive industry, providing detailed insights into the entire flow field and allowing engineers to simulate real-world open-road conditions with a fidelity that is often impractical or impossible to achieve through physical wind tunnel testing alone \cite{ashton2024drivaerml}.
Despite its utility, the application of CFD is fundamentally constrained by its immense computational expense. High-fidelity simulation methods, such as hybrid Reynolds-Averaged Navier-Stokes (RANS) - Large-Eddy Simulation (LES) or Direct Numerical Simulation (DNS), are necessary to accurately capture the complex, multi-scale turbulent phenomena inherent in automotive flows \cite{ashton2024drivaerml}. However, the computational cost of these methods scales unfavorably with the Reynolds number and the complexity of the vehicle geometry \cite{kochkov2021machine}. A typical high-fidelity simulation for a single design configuration can require hours or even days to run on large-scale high-performance computing (HPC) clusters, consuming thousands of core-hours \cite{ananthan2024machine}. This computational demand creates a significant bottleneck in the iterative design cycle, severely limiting the number of geometric variations that can be explored and hindering the pace of innovation \cite{choy2025factorized}.
In response to this grand challenge, the field of scientific machine learning has introduced the paradigm of data-driven surrogate models \cite{ref7}. These models, typically deep neural networks, are trained on large datasets of existing CFD simulations to learn a direct mapping from geometric parameters to aerodynamic solutions. Once trained, a surrogate model can produce predictions in seconds or milliseconds, offering a potential speed-up of several orders of magnitude compared to traditional solvers \cite{bleeker2025neuralcfd}. This dramatic acceleration promises to revolutionize the automotive design process, enabling near-real-time aerodynamic feedback, facilitating comprehensive design space exploration, and ultimately leading to more optimized and efficient vehicles \cite{ananthan2024machine}.

\subsection{The Proliferation and Specialization of Surrogate Architectures}

The application of machine learning to CFD has catalyzed a period of rapid architectural innovation, resulting in a diverse ecosystem of specialized models \cite{wang2024recent}. The research community has not yet converged on a single, universally optimal architecture; instead, different approaches have emerged, each with unique inductive biases that make them particularly well-suited for different facets of physics simulation. This proliferation reflects the complexity of the underlying physical phenomena and the various ways in which spatial and temporal dependencies can be represented and learned.
Three dominant classes of architectures have shown particular promise in the context of external aerodynamics. First, Graph Neural Networks (GNNs) have gained traction due to their natural ability to operate on the unstructured meshes commonly used in CFD \cite{ref11}. By representing mesh nodes as vertices and their connections as edges in a graph, GNNs use a message-passing mechanism to learn and propagate local physical interactions, making them highly effective for problems with complex, irregular geometries \cite{ref11}. Second, Neural Operators, particularly those based on Fourier transforms or point-cloud representations, have been developed to learn mappings between infinite-dimensional function spaces \cite{kong2025reducing}. These models aim to learn the solution operator of a partial differential equation itself, offering the potential for mesh-independent generalization. Finally, Convolutional Neural Networks (CNNs), which have dominated the field of computer vision, have been adapted for physics problems by projecting unstructured data onto regular grids \cite{cao2023efficient}. Their strengths lie in efficient feature extraction through hierarchical, translation-equivariant kernels, though they can face challenges with complex boundaries \cite{choy2025factorized}.
This architectural diversity leads to a crucial observation, often summarized by the "no free lunch" theorem in machine learning: there is no single model that is optimal for all problems. An architecture with a strong inductive bias for local interactions, like a GNN, may excel at resolving flow features around intricate geometric details like side mirrors or wheel wells. In contrast, an architecture with a more global receptive field, like a Fourier Neural Operator, might be superior at capturing large-scale phenomena such as the pressure distribution over the vehicle's body or the structure of the far-wake. This specialization implies that any single monolithic model, regardless of its sophistication, will likely exhibit a performance trade-off, excelling in some regions of the flow field while struggling in others where different physical principles are dominant.

\subsection{Proposed Approach: A Dynamic Ensemble of Heterogeneous Experts}

This work is founded on the hypothesis that the specialization of modern surrogate architectures should be viewed not as a limitation, but as an opportunity for synergistic combination. Rather than seeking a single "best" model, a more robust and accurate prediction can be achieved by creating a composite system that leverages the complementary strengths of multiple, diverse models. To this end, we propose a dynamic ensemble framework based on the Mixture of Experts (MoE) architecture \cite{jordan1994hierarchical}.
The MoE paradigm provides a sophisticated mechanism for model combination that goes far beyond simple techniques like prediction averaging. At its core, an MoE system consists of two key components: a set of "expert" networks, each a capable model in its own right, and a "gating network" that acts as a learned, dynamic router \cite{ref17}. For any given input, the gating network assesses the problem and produces a set of weights that determine the contribution of each expert to the final output. This process is input-conditional, meaning the gating network can learn to trust one expert for a specific region of the input space (e.g., the front of the car) and a different expert for another region (e.g., the rear spoiler) \cite{jordan1994hierarchical}. This allows the ensemble to adapt its strategy on a point-by-point basis, effectively assembling a specialized solution tailored to the local characteristics of the problem.
In this study, we construct an MoE system that combines three distinct, state-of-the-art surrogate models, chosen specifically for their architectural heterogeneity, which we believe is a key strength of the experimental design. The selected experts are: DoMINO, a point-cloud-based Decomposable Multi-scale Iterative Neural Operator. X-MeshGraphNet, a Scalable Multi-Scale Graph Neural Network. FigConvNet, a Factorized Implicit Global Convolution Network. By tasking a gating network with learning how to optimally combine the predictions from these three fundamentally different architectural paradigms, we aim to create a meta-model that is more accurate and robust than any of its individual components.

\subsection{Summary of Contributions}

This paper presents a significant advancement in the application of machine learning to industrial-scale engineering simulation. The primary contributions of this work are fourfold:
\begin{enumerate}
    \item \textbf{Novel Gating Network Design for Aerodynamics:} We design, implement, and train a novel Mixture of Experts gating network architecture specifically tailored for combining predictions of complex, high-dimensional surface fields—namely surface pressure and wall shear stress—in the challenging domain of external automotive aerodynamics.
    \item \textbf{First-of-its-Kind Heterogeneous Ensemble:} We provide the first demonstration of ensembling three architecturally distinct, state-of-the-art surrogate models for a large-scale, industrial-grade CFD problem. The deliberate choice of experts spanning point-cloud, graph-based, and convolutional paradigms represents a rigorous test of the MoE framework's ability to manage and leverage diverse inductive biases.
    \item \textbf{Application and Analysis of Entropy Regularization:} We systematically apply and analyze the impact of entropy regularization within the MoE loss function. This technique is shown to be crucial for promoting expert diversity, preventing the common failure mode of "expert collapse," and ensuring the gating network learns robust, physically plausible weighting strategies.
    \item \textbf{Comprehensive Validation on a High-Fidelity Benchmark:} We conduct a comprehensive quantitative and qualitative validation of the proposed MoE system on the DrivAerML benchmark dataset. The results provide definitive evidence that the MoE model achieves quantitatively superior performance, measured by L-2 relative error, compared to all three individual expert models across all predicted physical quantities.
\end{enumerate}
The research presented herein represents a strategic shift in the pursuit of ML-based surrogates. It moves beyond the search for a single superior architecture and toward the development of intelligent, adaptive systems that can orchestrate a portfolio of specialized models. This meta-learning approach provides a clear and extensible path toward building more powerful, generalizable, and reliable AI tools for scientific and engineering discovery.

\section{Background and Related Work}

\subsection{Machine Learning for Computational Fluid Dynamics}

The integration of machine learning into the CFD workflow has become a vibrant and rapidly expanding field of research. Recent comprehensive surveys have categorized the various roles ML can play into three main classes: Data-driven Surrogates, Physics-Informed Surrogates, and ML-assisted Numerical Solutions \cite{wang2024recent}. Data-driven surrogates, the category to which the present work belongs, aim to learn a direct mapping from a set of input parameters (such as geometry or boundary conditions) to the corresponding flow-field solution. This is achieved by training a deep neural network on a large database of pre-computed high-fidelity simulations. The primary goal is to create a model that, once trained, can perform inference at a fraction of the cost of the original numerical solver, thereby enabling rapid design exploration and optimization \cite{ananthan2024machine}.
Physics-Informed Neural Networks (PINNs) represent a second major category. Instead of relying solely on data, PINNs incorporate physical knowledge by embedding the governing partial differential equations (PDEs), such as the Navier-Stokes equations, directly into the training loss function as a form of soft constraint \cite{ref20}. This allows them to be trained with significantly less data and can help ensure that their predictions are physically consistent. The third category, ML-assisted numerical solutions, involves hybrid approaches where ML models are used to augment or accelerate specific components within a traditional CFD solver. This can include learning more accurate turbulence closure models, developing improved subgrid-scale models for LES, or accelerating the convergence of iterative linear solvers \cite{kochkov2021machine}.
The development of data-driven surrogates for industrial applications like automotive aerodynamics faces several key challenges that have driven architectural innovation. These include scalability, as models must be able to process geometries represented by millions of points or mesh elements; geometric complexity, requiring models that can handle the intricate and varied shapes of real-world vehicles; and generalization, the ability to make accurate predictions for vehicle shapes that were not seen during training \cite{kochkov2021machine}. The expert models selected for this study each represent a state-of-the-art approach to addressing these challenges through distinct architectural philosophies.

\subsection{A Survey of Modern Surrogate Architectures (The Experts)}

The success of a Mixture of Experts model is contingent upon the quality and diversity of its constituent experts. This work leverages three powerful, recently developed surrogate models, each representing a different fundamental approach to learning physics on geometric data.

\subsubsection{DoMINO: Decomposable Multi-scale Iterative Neural Operator}

DoMINO (Decomposable Multi-scale Iterative Neural Operator) is a novel architecture designed to address the challenges of scalability and geometric flexibility by operating directly on point-cloud representations of geometries \cite{ranade2025domino}. This approach is fundamentally mesh-free, which confers a significant practical advantage by eliminating the often time-consuming and labor-intensive step of mesh generation during the inference phase \cite{ranade2025domino}.
The architecture of DoMINO is built upon several key principles. It employs a multi-scale, iterative process to learn a global geometry encoding from the input point cloud \cite{ranade2025domino}. This global representation is designed to capture both short-range and long-range dependencies, which are crucial for accurately modeling the elliptic nature of the pressure field in aerodynamic flows \cite{ranade2025domino}. To enrich this encoding, the model incorporates additional geometric information, such as a Signed Distance Field (SDF) and positional encodings \cite{ranade2025domino}. The core predictive mechanism of DoMINO is local. For any point where a solution is desired, a dynamic computational stencil is constructed by sampling neighboring points \cite{ranade2025domino}. A local geometry encoding is extracted from the global representation, and this local information, combined with the stencil, is processed by a basis function neural network to predict the flow quantities at that point \cite{ranade2025domino}. This decomposable, local-to-global approach allows DoMINO to handle large-scale problems efficiently while maintaining high accuracy and generalizability to new geometries \cite{ranade2025domino}. The model has been successfully validated on the DrivAerML dataset for predicting both surface and volume flow fields \cite{ranade2025domino}.

\subsubsection{X-MeshGraphNet: Scalable Multi-Scale Graph Neural Networks}

Graph Neural Networks (GNNs) are a natural architectural choice for physics problems defined on unstructured meshes, as the mesh itself can be directly interpreted as a graph \cite{ref11}. Foundational models like MeshGraphNet established a powerful paradigm based on an encoder-processor-decoder structure, where the processor consists of multiple layers of message-passing, allowing information to propagate across the graph and simulate local physical interactions \cite{ref26}. However, early GNNs faced significant limitations in scalability to large industrial-sized meshes and often required a pre-existing simulation mesh for inference \cite{ref28}.
X-MeshGraphNet was developed as a scalable, multi-scale extension of MeshGraphNet to directly address these challenges \cite{nabian2024x}. Its first key innovation is a solution to the scalability bottleneck. It partitions a large input graph into smaller, manageable subgraphs that can be processed in parallel, for instance, across multiple GPUs \cite{nabian2024x}. To ensure that information flows seamlessly between these partitions, "halo" regions are created, which contain copies of nodes from adjacent subgraphs. This, combined with a gradient aggregation mechanism, ensures that training on the partitioned graph is mathematically equivalent to training on the full graph, but with drastically reduced memory requirements \cite{nabian2024x}. The second major innovation is the removal of the dependency on a simulation mesh. X-MeshGraphNet can construct its own custom, multi-scale graphs directly from standard CAD geometry files (e.g., STLs) \cite{nabian2024x}. It does this by generating a point cloud on the geometry's surface and connecting k-nearest neighbors to form the graph edges \cite{nabian2024x}. The multi-scale hierarchy is built by iteratively refining coarse point clouds to create finer ones, allowing the model to efficiently capture both local details and long-range interactions across the domain \cite{fortunato2022multiscale}. This combination of scalability and mesh-independence makes X-MeshGraphNet a highly practical and powerful tool for real-time simulation \cite{nabian2024x}.

\subsubsection{FigConvNet: Factorized Implicit Global Convolution Network}

Convolutional Neural Networks (CNNs) have proven to be exceptionally effective for tasks on regular grids, but their application to 3D physics problems has been hampered by computational complexity. Standard 3D convolutions have a computational and memory cost that scales cubically ($O(N^3)$) with the grid resolution N, which becomes prohibitive for the high-resolution domains required in industrial CFD \cite{choy2025factorized}. FigConvNet (Factorized Implicit Global Convolution Network) introduces a novel architectural solution to break this cubic scaling barrier \cite{choy2025factorized}.
The core methodology of FigConvNet is the factorization of the computational domain \cite{choy2025factorized}. Instead of representing a high-resolution 3D domain (e.g., $1024 \times 1024 \times 1024$) with a single dense grid, FigConvNet approximates it using a set of "Factorized Implicit Grids." Each grid in this set is high-resolution in two dimensions but has a very low resolution in the third (e.g., $1024 \times 1024 \times 4$, $1024 \times 4 \times 1024$, and $4 \times 1024 \times 1024$) \cite{choy2025factorized}. This decomposition dramatically reduces the total number of elements that need to be stored and processed. The model then approximates a full 3D convolution by performing more efficient 2D global convolutions on each of these planes in parallel, reducing the overall complexity to quadratic ($O(N^2)$) \cite{choy2025factorized}. To ensure that information is integrated across these separate factorized representations, a data fusion mechanism using trilinear interpolation is employed \cite{choy2025factorized}. This entire factorized convolution block is integrated within a U-shaped network architecture (U-Net), which is highly effective at capturing multi-scale features and preserving high-resolution details through skip connections, enabling accurate prediction of both global quantities like drag and detailed surface pressure fields \cite{choy2025factorized}.
The selection of these three experts—one point-based, one graph-based, and one grid-based—provides a rich and diverse foundation for the MoE framework. Each model embodies a distinct philosophy for discretizing and learning physical systems, analogous to the classical numerical methods of meshless, finite element, and finite difference schemes, respectively. The gating network is therefore tasked not merely with combining model outputs, but with learning to arbitrate between these fundamental computational paradigms, selecting the most appropriate approach for the local physics at every point on the vehicle's surface.

\subsection{Ensemble Learning in Aerodynamic Modeling}

The concept of combining multiple models to produce a single, superior prediction—known as ensemble learning—is a well-established principle in machine learning and has seen application in aerodynamic modeling \cite{cheng2025integrated}. Traditional ensembling techniques often involve simple strategies like averaging the predictions of several independently trained models. This approach can improve robustness and reduce variance, but it is static and does not account for the possibility that different models may have varying levels of expertise across the problem domain \cite{petrov2024application}. More advanced methods like stacking involve training a second-level meta-model to learn a weighted combination of the base models' predictions, but these weights are typically global and not input-dependent \cite{wang2024airfoil}.
The Mixture of Experts (MoE) framework represents a significant conceptual leap beyond these methods \cite{jordan1994hierarchical}. As a "divide and conquer" algorithm, MoE partitions the problem space and assigns different experts to each sub-problem \cite{mu2025comprehensive}. The key component is the gating network, which takes the same input as the experts and outputs a probability distribution over them. This distribution determines how the experts' outputs are combined to form the final prediction. In modern deep learning applications, particularly in large language models, this is often implemented as a sparse routing mechanism, where only the top-k experts with the highest gating scores are activated for a given input token, leading to massive computational savings \cite{ref17}.
In the context of this work, we employ a "soft" MoE, where the gating network outputs a dense weighting vector, and the final prediction is a weighted sum of all expert outputs. The critical feature that distinguishes this from simpler ensembles is that the gating weights are computed dynamically for each and every point in the input space. This allows the model to learn a highly granular, spatially-variant strategy for leveraging expert knowledge. While ensemble methods have been explored in aerodynamics, the novelty of the present work lies in two key aspects: first, the use of the dynamic, conditional computation inherent to the MoE framework, which is far more sophisticated than static averaging or global weighting schemes. Second, the unprecedented complexity and architectural diversity of the models employed as experts. Combining state-of-the-art models like DoMINO, X-MeshGraphNet, and FigConvNet within an MoE framework for a problem of this scale and complexity represents a novel and significant contribution to the field of scientific machine learning.

\section{A Gating Network for Aerodynamic Ensembling}

The core of our proposed methodology is a Mixture of Experts (MoE) system designed to synergistically combine the predictions of multiple pre-trained aerodynamic surrogate models. This section details the architecture of the system, with a particular focus on the design of the gating network and the formulation of the training objective, which includes a crucial entropy regularization term.

\subsection{System Overview}

The overall system operates as a two-stage predictive pipeline. In the first stage, the three independent "expert" models—DoMINO, X-MeshGraphNet, and FigConvNet—are queried in parallel. Each expert takes the vehicle geometry as input and produces a full-field prediction for surface pressure (P) and the wall shear stress vector (WSS) across all points of the surface mesh. These three sets of predictions, which represent the specialized "opinions" of each expert, are then passed to the second stage.
The second stage consists of the MoE Gating Network. This network does not see the original geometry directly; instead, its inputs are the predictions generated by the experts, along with local geometric features such as surface normals. The gating network's function is to act as a learned arbiter, analyzing the expert predictions at each point on the vehicle surface and determining the optimal weighting for combining them. It outputs a set of weights for pressure and another set for wall shear stress, which are then used to compute the final, refined MoE prediction. This architecture effectively creates a meta-learning system where the gating network learns about the behavior and relative strengths of the expert models.

\subsection{MoE Gating Network Architecture}

The gating network is implemented as a deep neural network that processes information on a point-wise basis. A crucial design choice, informed by the underlying physics, is the use of two separate and independent gating modules: one dedicated to predicting the weights for the scalar pressure field, and another for the vector wall shear stress field. This separation acknowledges that the physical phenomena governing pressure (largely inviscid, related to global curvature and flow acceleration) are distinct from those governing wall shear stress (viscous, related to the boundary layer and local friction). A model that excels at predicting one quantity may not be the best at predicting the other, and separate gating networks allow the system to learn these distinct expertise criteria independently.
\textbf{Input Features:}
For each point on the surface mesh, a feature vector is constructed and fed into the gating networks. This vector is formed by concatenating the following information:
\begin{itemize}
    \item \textbf{Expert Pressure Predictions:} The scalar pressure values predicted by each of the three experts: $P_{\text{DoMINO}}$, $P_{\text{FigConvNet}}$, and $P_{\text{XMGN}}$.
    \item \textbf{Expert Shear Stress Predictions:} The 3-component wall shear stress vectors predicted by each expert: $\text{WSS}_{\text{DoMINO}}$, $\text{WSS}_{\text{FigConvNet}}$, and $\text{WSS}_{\text{XMGN}}$.
    \item \textbf{Local Geometric Features:} To provide the network with context about the local surface geometry, the 3-component surface normal vector ($n_x, n_y, n_z$) is optionally included.
\end{itemize}

\textbf{Network Structure:}
Both the pressure and wall shear stress gating modules are implemented as Multi-Layer Perceptrons (MLPs). Based on the provided configuration, each MLP consists of 3 hidden layers, with each layer containing 128 neurons. The Rectified Linear Unit (ReLU) is used as the activation function for all hidden layers.

\textbf{Output Layer and Weight Generation:}
The final layer of each MLP produces a vector of logits with a dimension equal to the number of experts (in this case, 3). A softmax activation function is then applied to these logits to transform them into a normalized probability distribution, representing the weights for each expert. For the pressure prediction at a given point, the gating network outputs weights ($W_{p,1}, W_{p,2}, W_{p,3}$) such that $\sum_{i=1}^{3} W_{p,i} = 1$. Similarly, the shear stress gating network outputs weights ($W_{s,1}, W_{s,2}, W_{s,3}$).
The system can also be configured to correct for systematic biases, where all experts might consistently over- or under-predict in certain regions. When this bias correction is enabled, the gating network's final layer is augmented to output an additional value, which serves as an additive correction term, $C_{\text{MoE}}$.

\textbf{Mathematical Formulation:}
The final MoE prediction for pressure, $P_{\text{MoE}}$, at a single point is computed as a weighted sum of the expert predictions, optionally including the learned bias term:

\begin{equation}
P_{\text{MoE}} = W_{p,1} \cdot P_{\text{DoMINO}} + W_{p,2} \cdot P_{\text{FigConvNet}} + W_{p,3} \cdot P_{\text{XMGN}} + C_{p, \text{MoE}}
\end{equation}

A corresponding equation is used for each component of the wall shear stress vector, $\text{WSS}_{\text{MoE}}$:

\begin{equation}
\text{WSS}_{\text{MoE}} = W_{s,1} \cdot \text{WSS}_{\text{DoMINO}} + W_{s,2} \cdot \text{WSS}_{\text{FigConvNet}} + W_{s,3} \cdot \text{WSS}_{\text{XMGN}} + C_{s, \text{MoE}}
\end{equation}

The weights ($W_{p,i}, W_{s,i}$) and the correction terms ($C_{p, \text{MoE}}, C_{s, \text{MoE}}$) are all outputs of the gating network, dynamically computed for each point on the mesh.

\subsection{Training with Entropy Regularization}

A common challenge in training MoE models is the tendency for the gating network to converge to a trivial solution where it relies exclusively on a single expert, a phenomenon known as "expert collapse" or "model collapse." This can happen if one expert is marginally better than the others across the majority of the training data. The gating network quickly learns to assign it a weight of nearly 1, effectively ignoring the other experts. This outcome is undesirable as it fails to leverage the complementary knowledge that the other experts might possess, defeating the purpose of the ensemble.
To counteract this, we introduce an entropy regularization term into the training loss function. The Shannon entropy of the gating network's weight distribution, $w = (w_1, w_2, \dots, w_N)$, is given by:

\begin{equation}
H(w) = - \sum_{i=1}^{N} w_i \log(w_i)
\end{equation}

Entropy is a measure of uncertainty or "flatness" of a probability distribution. A distribution with high entropy is close to uniform (e.g., weights of [0.33, 0.33, 0.33]), while a distribution with low entropy is sharply peaked (e.g., weights of [0.98, 0.01, 0.01]). By adding a term to the loss function that encourages the maximization of entropy, we penalize the gating network for being over-confident in any single expert. This forces the network to maintain a more balanced distribution of weights, ensuring that all experts are considered and have an opportunity to contribute to the final prediction.

\textbf{Loss Function Formulation:}
The total loss function, $L_{\text{total}}$, is a composite of the primary prediction losses and the entropy regularization terms. The prediction losses for pressure and shear stress are calculated as the Mean Squared Error (MSE) between the MoE predictions and the ground truth data:

\begin{equation}
L_{\text{pressure}} = \text{MSE}(P_{\text{MoE}}, P_{\text{true}})
\end{equation}

\begin{equation}
L_{\text{shear}} = \text{MSE}(\text{WSS}_{\text{MoE}}, \text{WSS}_{\text{true}})
\end{equation}

The total loss is then formulated to minimize the prediction error while maximizing the entropy of the weight distributions for both pressure ($w_{\text{pressure}}$) and shear stress ($w_{\text{shear}}$). Since standard optimization minimizes the loss, maximizing entropy is equivalent to minimizing its negative. The final loss function is:

\begin{equation}
L_{\text{total}} = L_{\text{pressure}} + L_{\text{shear}} - \lambda_{\text{entropy}} \cdot (H(w_{\text{pressure}}) + H(w_{\text{shear}}))
\end{equation}

The hyperparameter $\lambda_{\text{entropy}}$ is a non-negative coefficient that controls the strength of the regularization. A value of $\lambda_{\text{entropy}} = 0$ would disable the regularization entirely, while a larger value would more strongly encourage expert diversity. This regularization is critical for training a robust gating network that produces reliable scores and effectively harnesses the collective intelligence of the expert ensemble.

\section{Experimental Design}

To rigorously evaluate the performance of the proposed Mixture of Experts gating network, we designed an experimental protocol centered on a large-scale, high-fidelity public benchmark dataset. This section details the dataset, the implementation and training procedures, and the metrics used for evaluation.

\subsection{The DrivAerML Benchmark Dataset}

The foundation of this study is the DrivAerML dataset, a comprehensive, open-source (CC-BY-SA) public dataset created specifically to accelerate the development of machine learning models for automotive aerodynamics \cite{ashton2024drivaerml}. The lack of large, high-quality, publicly available training data has long been a barrier to progress in the field, and DrivAerML was conceived to directly address this gap \cite{ashton2024drivaerml}.

\textbf{Geometry:} The dataset is built upon the DrivAer generic vehicle model, a widely used and well-documented benchmark in both academia and industry \cite{ashton2024drivaerml}. It comprises 500 unique geometric variants derived from a parametric morphing of the baseline DrivAer notchback configuration \cite{ashton2024drivaerml}. This parametric variation ensures a diverse range of shapes, covering the main features and geometric trends seen in modern road vehicles, which is essential for training generalizable ML models \cite{ashton2024drivaerml}.

\textbf{CFD Fidelity:} A key distinguishing feature of DrivAerML is the exceptional fidelity of its simulation data. Unlike many other datasets that rely on steady-state RANS simulations, all 500 cases in DrivAerML were generated using a high-fidelity, scale-resolving CFD approach—specifically, a hybrid RANS-LES turbulence model \cite{ashton2024drivaerml}. This methodology is representative of the state-of-the-art workflows used within the automotive industry for simulations where high accuracy is paramount \cite{bleeker2025neuralcfd}. The simulations were executed using a customized version of the open-source CFD code OpenFOAM, following consistent and validated automated workflows to ensure the quality and consistency of the entire dataset \cite{ashton2024drivaerml}. This high-fidelity foundation is crucial, as the accuracy of any data-driven surrogate model is ultimately limited by the quality of the data on which it is trained.

\textbf{Data Richness:} The dataset is exceptionally rich, providing a wealth of information for each of the 500 samples. The data is provided in open-source file formats to maximize accessibility. For each geometric variant, the dataset includes \cite{ashton2024drivaerml}:
\begin{itemize}
    \item The geometry itself in STL format.
    \item Boundary surface data containing predicted fields in VTK Polygonal Data format.
    \item Full volumetric flow field data in VTK Unstructured Grid format.
    \item Time-averaged aerodynamic forces and moment coefficients in CSV format.
    \item A collection of 2D slices of the volumetric flow field.
\end{itemize}
This comprehensive data structure supports a wide range of ML applications, from training surrogate models for surface quantities, as in this work, to generative design and full 3D flow field prediction \cite{nabian2024x}. For this study, the ground truth labels for surface pressure and wall shear stress are extracted from the files.

\subsection{Implementation and Training Protocol}

The training pipeline for the MoE gating network is implemented in NVIDIA PhysicsNeMo.

\textbf{Data Preprocessing:} The initial step involves a preprocessing script (\texttt{preprocessor.py}) that prepares the raw DrivAerML data for training. This script performs several key functions. First, it iterates through the dataset and loads the ground truth surface data for each sample. Concurrently, it loads the corresponding surface predictions generated by each of the three expert models (DoMINO, X-MeshGraphNet, and FigConvNet), which have been pre-computed for the entire dataset. It then combines all these fields—expert predictions, ground truth labels, and geometric features like surface normals—into a single, unified data structure. The script calculates normalization statistics (mean and standard deviation) for all input and output fields across the training set, which are essential for stable network training. Finally, it splits the dataset into training and validation sets and saves the fully processed data as a new series of \texttt{.vtp} files, which are then consumed by the data loader during training.

\textbf{Training Configuration:} All hyperparameters for the model architecture and training process are managed through a central \texttt{config.yaml} file, ensuring reproducibility. The model is trained with a \texttt{batch\_size} of 1 (as each sample contains a large point cloud representing a full vehicle). Training is conducted for \texttt{num\_epochs: 10}. A learning rate scheduler is employed, starting with an initial learning rate of \texttt{start\_lr: 1e-3} and decaying to a final value of \texttt{end\_lr: 5e-6}, likely following a cosine annealing schedule to facilitate fine-tuning in the later stages of training. The entropy regularization weight is set to \texttt{lambda\_entropy: 0.01}.

\textbf{Computational Framework:} The training process is optimized for performance on modern GPU hardware. The framework supports distributed data-parallel training using \texttt{torchrun}, allowing the workload to be efficiently scaled across multiple GPUs to reduce training time. To further accelerate computation and reduce memory usage, automatic mixed precision (AMP) training is enabled. This technique performs certain operations in 16-bit floating-point format instead of the standard 32-bit, while maintaining model accuracy. Throughout the training process, key metrics such as the pressure loss, shear stress loss, total loss, entropy, and the learning rate are logged to TensorBoard for real-time monitoring and post-hoc analysis.

\subsection{Evaluation Metrics and Inference}

\textbf{Evaluation Metric:} The primary quantitative metric used to evaluate the performance of the MoE model and the individual experts is the L-2 relative error norm. For a predicted field $u_{\text{pred}}$ and a ground truth field $u_{\text{true}}$, each defined over N points, the L-2 relative error is calculated as:

\begin{equation}
\text{L-2 Error} = \frac{\sqrt{\sum_{i=1}^{N} (u_{\text{pred},i} - u_{\text{true},i})^2}}{\sqrt{\sum_{i=1}^{N} (u_{\text{true},i})^2}}
\end{equation}

This metric provides a normalized measure of the overall prediction error, making it suitable for comparing performance across different physical quantities and models.

\textbf{Inference Process:} After training is complete, the performance of the final model checkpoint is evaluated on the held-out test set using the \texttt{inference.py} script. This script loads the trained gating network, iterates through the test samples, and for each sample, generates the final MoE-weighted predictions for pressure and wall shear stress. The script outputs the results in \texttt{.vtp} file format. Crucially, these output files contain not only the final blended predictions but also the pointwise expert weights (scores) that were assigned by the gating network. This allows for detailed qualitative analysis and visualization of the gating network's decision-making process, enabling an understanding of why the MoE model performs as it does by revealing which experts it trusts in different regions of the vehicle. These visualizations are typically generated using scientific visualization software such as ParaView.

\section{Results and Discussion}

The efficacy of the Mixture of Experts framework is evaluated through a comprehensive analysis of both quantitative error metrics and qualitative visualizations of the gating network's learned behavior. The results collectively demonstrate that the MoE model not only outperforms its constituent experts but also learns a physically meaningful strategy for combining their predictions.

\subsection{Quantitative Superiority of the MoE Model}

The primary objective of this work is to develop a composite model that achieves higher predictive accuracy than any single expert model. The quantitative results, summarized in Table 1, provide clear and compelling evidence that this objective has been met. The table presents a comparison of the L-2 relative errors for the MoE model against each of the three individual expert models—DoMINO, X-MeshGraphNet, and FigConvNet—across the test set of the DrivAerML dataset.

\begin{table}[ht]
\centering
\caption{Comparative L-2 relative error analysis of the Mixture of Experts (MoE) model and the three individual expert models for surface pressure (P) and the three components of wall shear stress (WSS). The MoE model demonstrates the lowest error across all predicted quantities.}
\label{tab:comparison}
\footnotesize
\begin{tabular}{l|rrrr}
\textbf{Model} & \textbf{P L-2 Error} & \textbf{WSS (x) L-2 Error} & \textbf{WSS (y) L-2 Error} & \textbf{WSS (z) L-2 Error} \\\hline
MoE & 0.08 & 0.14 & 0.19 & 0.21 \\
DoMINO & 0.10 & 0.18 & 0.26 & 0.28 \\
X-MeshGraphNet & 0.14 & 0.17 & 0.22 & 0.29 \\
FigConvNet & 0.21 & 0.32 & 0.62 & 0.53
\end{tabular}
\end{table}

The results in Table \ref{tab:comparison} unequivocally establish the superior performance of the MoE model. It achieves the lowest L-2 error for the pressure prediction and for all three components of the wall shear stress vector. This consistent improvement across all measured quantities highlights the robustness of the learned ensembling strategy.

A deeper analysis of the results reveals several important points. First, the performance improvement is significant. For pressure prediction, the MoE model achieves an L-2 error of 0.08, which represents a 20\% relative error reduction compared to the best-performing individual expert, DoMINO (0.10), and a remarkable 61.9\% reduction compared to the weakest expert, FigConvNet (0.21). Similar substantial gains are observed for the wall shear stress components. For instance, in the WSS (y) component, the MoE model's error of 0.19 is 13.6\% lower than the best expert for that component, X-MeshGraphNet (0.22).

Second, the table reveals a clear performance hierarchy among the individual experts. DoMINO emerges as the strongest single model, particularly for pressure prediction. X-MeshGraphNet is highly competitive, especially for wall shear stress, while FigConvNet exhibits the highest error across the board. This heterogeneity is not a weakness of the experiment but rather a crucial ingredient for the success of the MoE. The fact that the MoE model can substantially outperform even the strongest expert (DoMINO) provides definitive proof that the weaker experts, X-MeshGraphNet and FigConvNet, are not simply adding noise. Instead, they must be providing valuable, non-redundant information in specific regions or for specific flow phenomena that DoMINO fails to capture as accurately. The gating network's success lies in its ability to identify these localized "pockets of expertise" from all available models, even those that are globally weaker, and to leverage them to correct the errors of the strongest model. This demonstrates a powerful principle of intelligent ensembling: the final composite prediction is superior to the contribution of any single member.

\subsection{Analysis of Spatially-Variant Expert Contributions}

While the quantitative metrics in Table \ref{tab:comparison} confirm that the MoE model works, a qualitative analysis of the gating network's weights is necessary to understand how it works. By visualizing the pointwise weights assigned to each expert during inference, we can gain insight into the learned decision-making strategy. Figure \ref{fig:weights} illustrates these expert weights for the pressure prediction on a representative sample (ID 129) from the DrivAerML test set.

The visualization in Figure \ref{fig:weights} reveals a highly structured and physically interpretable weighting pattern. The gating network has not learned a simple or uniform weighting scheme; instead, it has developed a sophisticated, spatially-variant strategy that aligns remarkably well with the known architectural strengths and inductive biases of the expert models.
Several key regions of interest can be identified. At the very front of the vehicle, particularly at the stagnation point on the grille where the incoming flow is brought to rest, the gating network assigns a very high weight to DoMINO. This is a region of maximum pressure, and its behavior is governed by large-scale, global flow dynamics. DoMINO, as a neural operator designed to capture both short- and long-range interactions, is well-suited to model this phenomenon accurately.
In contrast, X-MeshGraphNet is favored in regions of high geometric complexity and sharp curvature. For example, it receives higher weight around the side mirrors, along the sharp A-pillars framing the windshield, and near the complex junctions of the rear spoiler. As a Graph Neural Network, X-MeshGraphNet's core strength lies in its ability to model local interactions on unstructured meshes, making it an ideal expert for capturing the rapid changes in pressure that occur around such intricate geometric features.
Finally, FigConvNet is assigned significant weight in specific areas, primarily the large, relatively smooth, and flat surfaces such as the roof, the doors, and the underbody of the vehicle. These regions are more amenable to the grid-based feature extraction performed by convolutional architectures.
This qualitative analysis provides powerful evidence that the gating network is performing a form of learned, physics-informed model selection at the local level. It has successfully identified the distinct strengths of the point-based, graph-based, and grid-based architectural paradigms and has learned to deploy each one precisely where its inductive bias is most beneficial. This demonstrates that the MoE framework is not just a black-box blender but a sophisticated meta-learning system capable of reasoning about the suitability of different models for different physical and geometric contexts.

\subsection{The Role of Entropy Regularization}

The physically plausible and spatially diverse weighting patterns shown in Figure \ref{fig:weights} are a direct result of applying entropy regularization during training. This component of the loss function is essential for preventing the gating network from collapsing into a state of over-confidence in a single expert.

\begin{figure}[H]
\centering
\includegraphics[width=\textwidth]{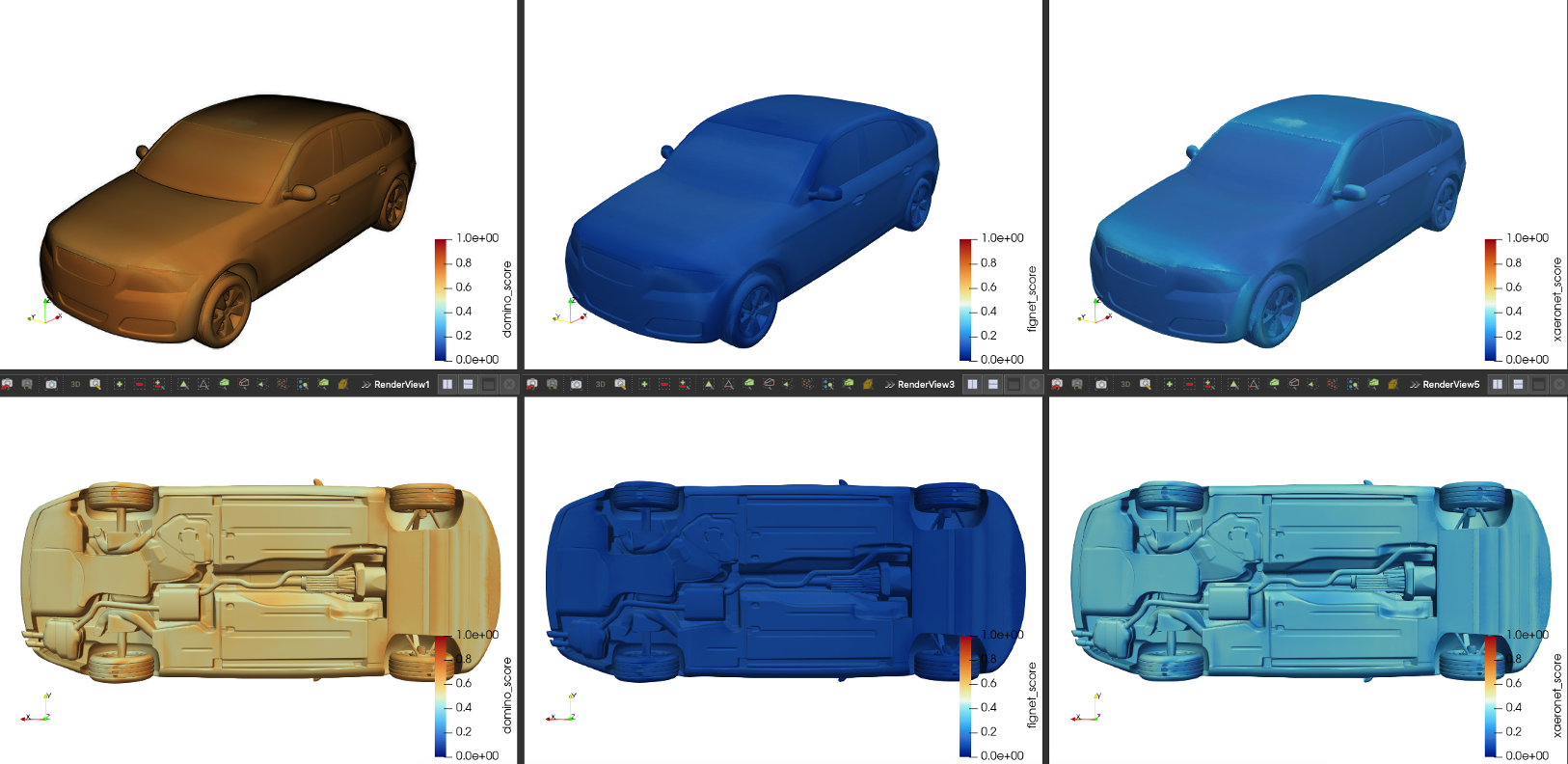} 
\caption{Demonstration of model collapse when training without entropy regularization. For pressure prediction on a DrivAerML sample, the gating network assigns nearly all weight to a single expert (DoMINO, shown in blue) across the entire vehicle surface. This "winner-take-all" behavior fails to leverage the strengths of the other experts.}
\label{fig:no_regularization}
\end{figure}

To demonstrate its importance, we conducted an ablation study. Without regularization, the gating network learns to predominantly favor the strongest individual expert, DoMINO, across nearly the entire vehicle surface (Figure \ref{fig:no_regularization}). This "winner-take-all" outcome defeats the purpose of the MoE framework.

\begin{figure}[H]
\centering
\includegraphics[width=\textwidth]{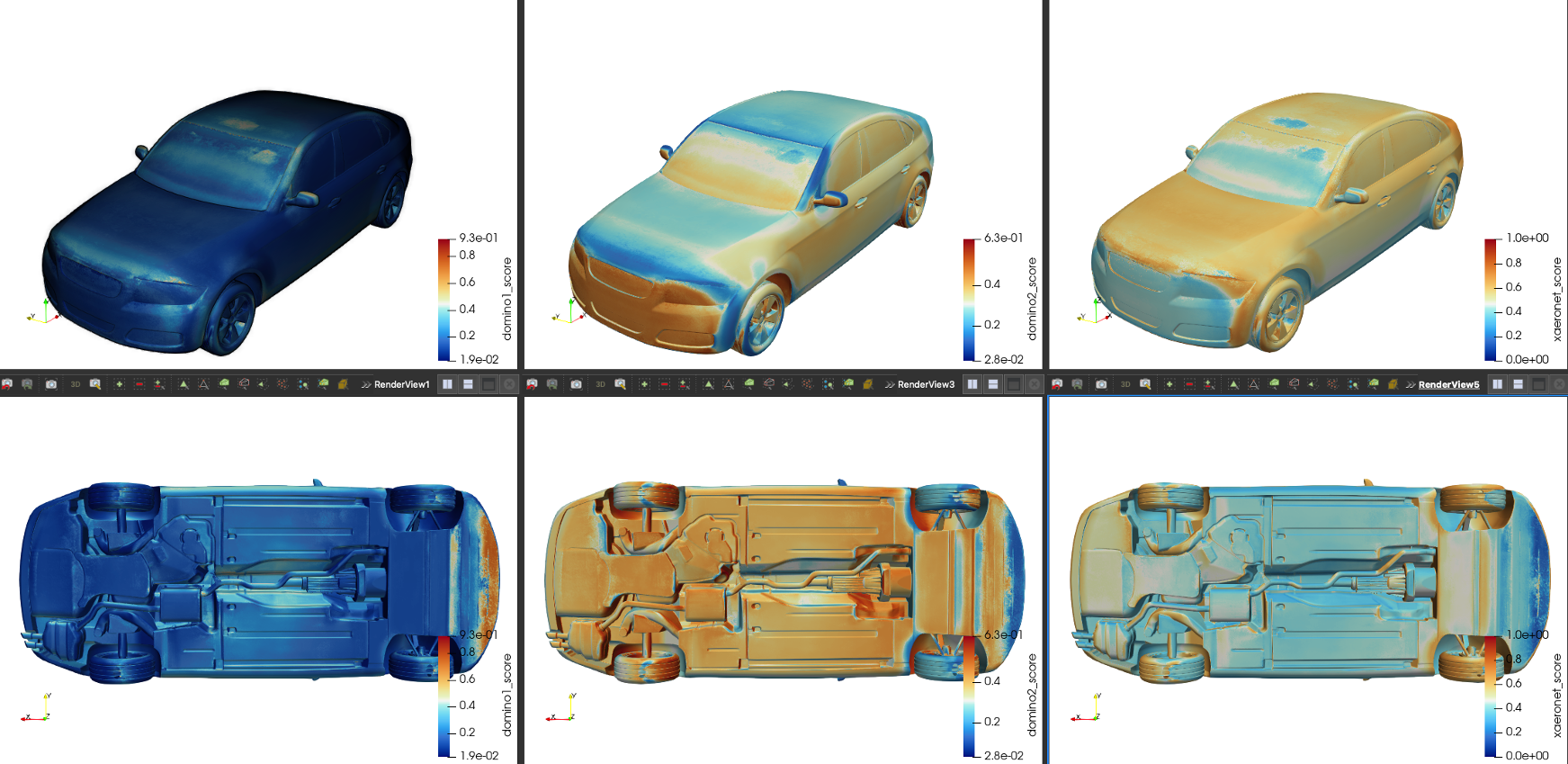} 
\caption{Unreliable expert scoring without entropy regularization. Duplicating DoMINO predictions as a second input reveals that the gating network assigns inconsistent and arbitrary weights to these identical experts, indicating a failure to provide robust scores.}
\label{fig:double_domino_input}
\end{figure}

To further underscore the instability of this unregulated model, we performed an experiment where the FigConvNet input was replaced with a duplicate of the DoMINO prediction. The gating network assigned vastly different weights to these two identical inputs (Figure \ref{fig:double_domino_input}), confirming that its scoring mechanism is arbitrary and unreliable without regularization.

\begin{figure}[H]
\centering
\includegraphics[width=\textwidth]{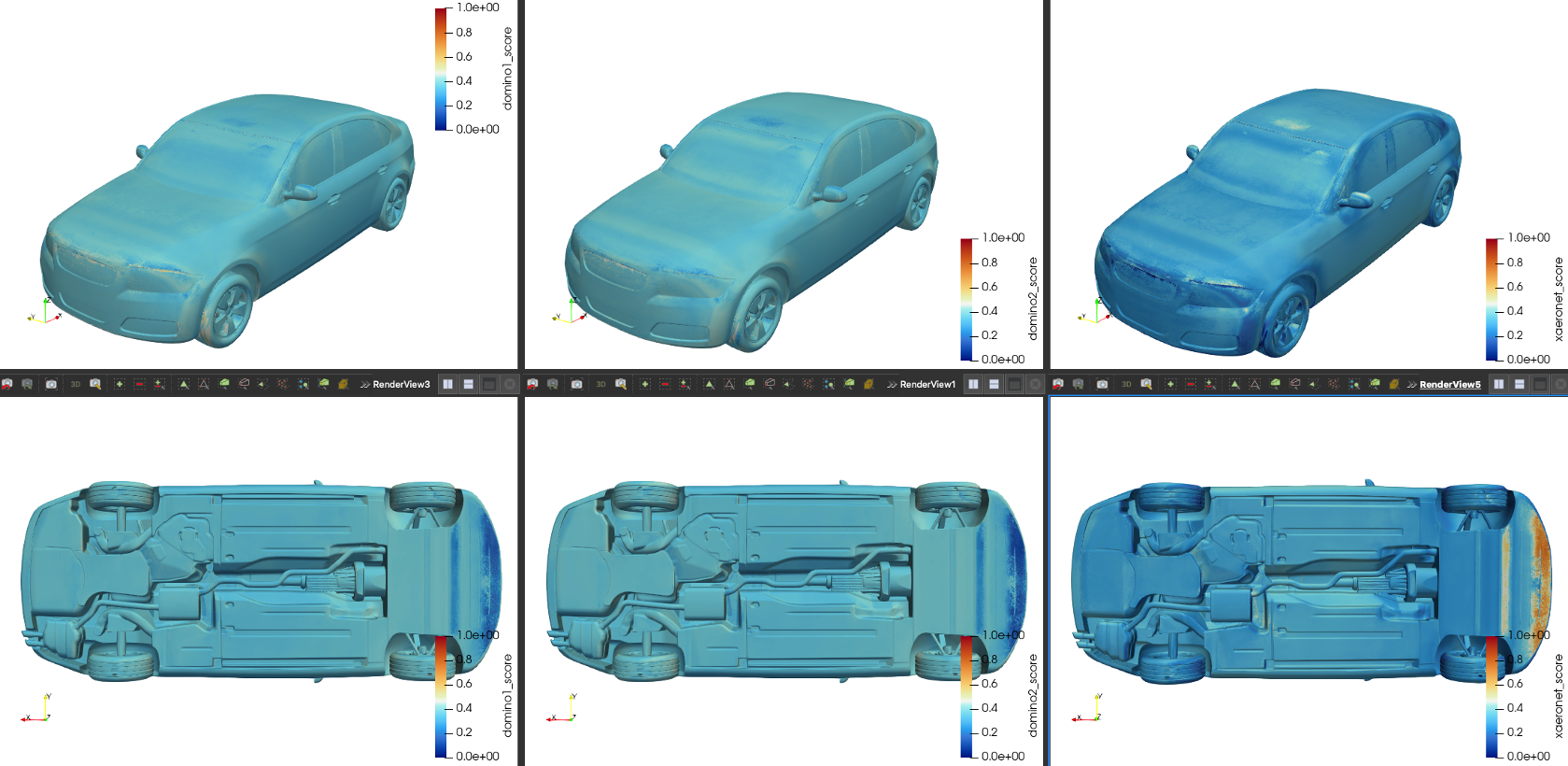} 
\caption{Demonstration of robust scoring with entropy regularization. When presented with two identical DoMINO expert inputs, the gating network correctly assigns them equal weight, validating the consistency of the regularized model.}
\label{fig:double_domino_max_entropy}
\end{figure}

With regularization enabled, the network is forced to explore more balanced weight distributions. This encouragement of a "soft" blending between experts prevents sharp, discontinuous transitions in the weighting field and contributes to a more robust final prediction. When we repeated the duplicated-input experiment with regularization active, the model correctly assigned identical weights to the two identical DoMINO models (Figure \ref{fig:double_domino_max_entropy}). This result validates that the regularized gating mechanism produces reliable, well-calibrated scores.

\begin{figure}[H]
\centering
\includegraphics[width=\textwidth]{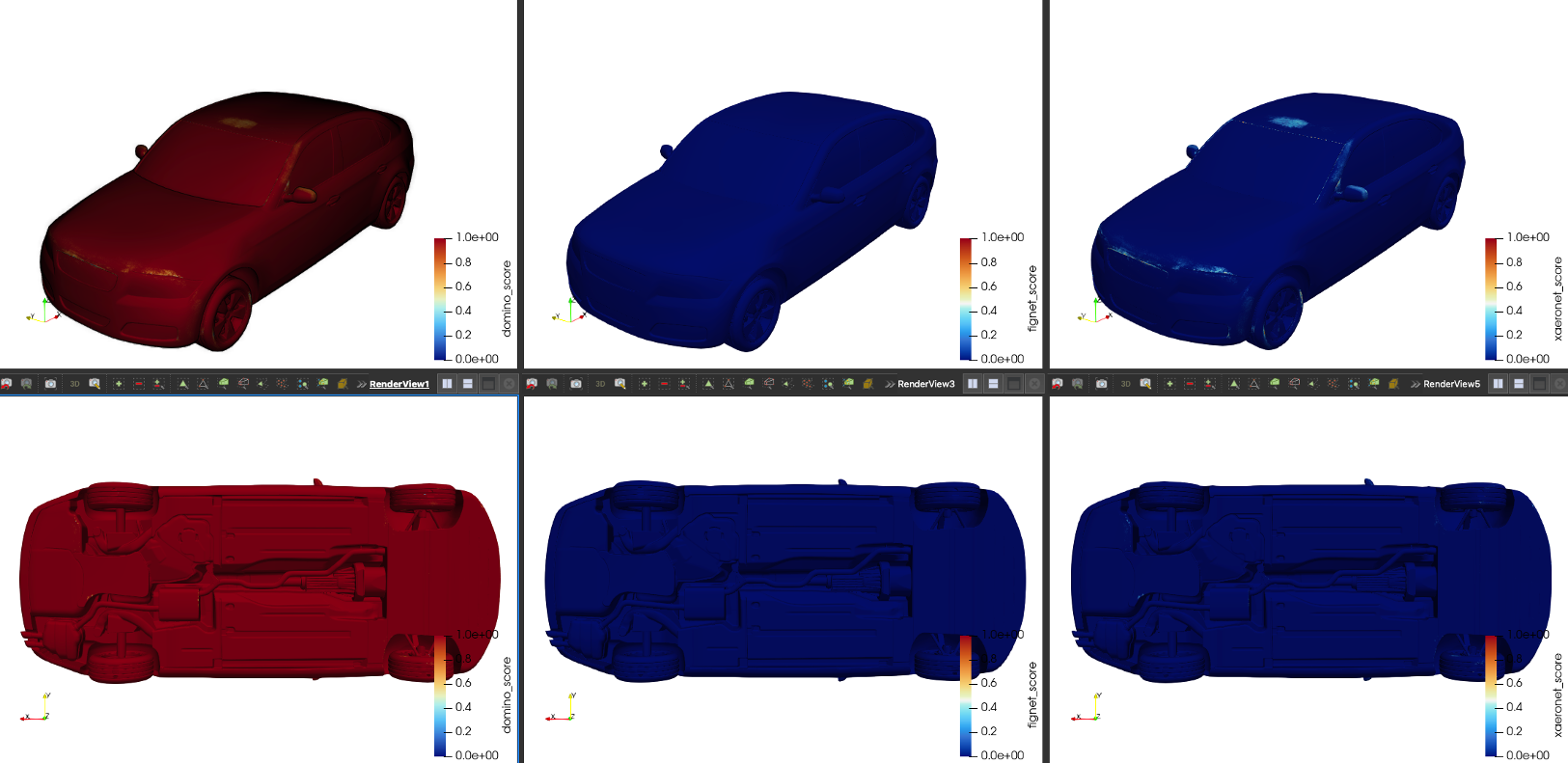} 
\caption{Gating weights learned when minimizing entropy. This objective function causes the model to become over-confident in a single expert (DoMINO), effectively ignoring X-MeshGraphNet and FigConvNet. This demonstrates the expected inverse behavior to entropy maximization.}
\label{fig:min_entropy}
\end{figure}

Finally, as a conceptual sanity check, we also investigated the effect of minimizing entropy. As expected, this objective guided the model to reduce uncertainty and confidently rely on a single expert (Figure \ref{fig:min_entropy}). These experiments confirm that entropy regularization is not merely a training heuristic but a fundamental component of our methodology, enabling the MoE framework to achieve its goal of synergistic expert collaboration.

\section{Conclusion}

This research has addressed the challenge of improving the accuracy of data-driven surrogate models for external automotive aerodynamics by proposing and validating a novel Mixture of Experts (MoE) framework. By moving beyond the paradigm of developing single monolithic architectures, this work has demonstrated that a dynamic, intelligent ensemble of heterogeneous experts can achieve a new level of predictive performance.

\subsection{Recapitulation of Findings}

The central finding of this study is that a dedicated gating network, trained with entropy regularization, can effectively learn to combine the predictions from a diverse set of state-of-the-art surrogate models. The proposed MoE system, which integrates a point-cloud-based neural operator (DoMINO), a scalable graph neural network (X-MeshGraphNet), and a factorized convolutional network (FigConvNet), was shown to be highly effective.
The quantitative evaluation, conducted on the high-fidelity DrivAerML benchmark dataset, provided conclusive evidence of the MoE model's superiority. It achieved significantly lower L-2 relative errors in predicting both surface pressure and wall shear stress compared to any of the individual expert models. This result is particularly noteworthy as the MoE model outperformed even the strongest individual expert, indicating that it successfully extracted and utilized valuable, complementary information from all members of the ensemble.
Furthermore, the qualitative analysis of the gating network's learned weights revealed a sophisticated, physically meaningful strategy. The network learned to assign credibility on a point-by-point basis, favoring different experts in different regions of the vehicle in a manner consistent with their underlying architectural strengths. This spatially-variant weighting scheme confirms that the MoE framework is not merely averaging predictions but is performing a nuanced, input-conditional model selection, leading to a composite prediction that is more accurate and robust than its parts.

\subsection{Broader Impact and Implications}

The success of this approach has significant implications for both industrial engineering practice and the broader field of scientific machine learning. For the automotive industry, the ability to create more accurate and reliable surrogate models can further accelerate the design-simulation-optimization loop. By providing designers with faster and more trustworthy aerodynamic feedback, this technology can facilitate more extensive design space exploration, leading to vehicles with improved fuel efficiency, stability, and safety.
More broadly, this study serves as a powerful proof-of-concept for a generalizable meta-learning strategy. The challenge of having multiple competing, specialized, and imperfect models is not unique to CFD but is common across many scientific domains where ML is being applied, including structural mechanics, thermal analysis, materials science, and climate modeling. The MoE framework presented here provides a robust and principled methodology for combining the strengths of diverse models in any of these fields. It suggests a path forward where the research community can focus not only on creating new expert architectures but also on developing more sophisticated meta-models that can intelligently orchestrate an ever-growing portfolio of specialized tools.

\subsection{Future Directions}

This work opens up several promising avenues for future research that can build upon the foundation established here.

\textbf{Expanding the Expert Pool:} The MoE framework is inherently modular. Future work should explore the integration of additional, even more diverse expert models. As new architectures are developed in the field, they can be readily incorporated into the ensemble, potentially leading to further accuracy gains. The process for adding new experts is straightforward, primarily involving updates to the data preprocessing and model configuration files.

\textbf{More Sophisticated Gating Mechanisms:} While the MLP-based gating network proved effective, more advanced architectures could be explored. For instance, a graph-based gating network could reason more globally about the vehicle's topology when assigning weights, or an attention-based mechanism could learn to explicitly focus on the most informative expert predictions when making its decisions.

\textbf{Extension to Volumetric Prediction:} The current work focused on predicting surface quantities. A natural and highly valuable extension would be to apply the MoE framework to the prediction of the full 3D volumetric flow field (velocity, pressure, and turbulence quantities) in the space surrounding the vehicle. This would require adapting the gating network to operate on 3D coordinates and would provide a more complete aerodynamic picture for analysis.

\textbf{Uncertainty Quantification:} The output of the gating network's softmax layer is a probability distribution. The entropy of this distribution could potentially serve as a valuable, built-in proxy for model uncertainty. Regions where the entropy is high (i.e., the gating network is uncertain which expert to trust) could be flagged as areas of low confidence in the prediction, providing a powerful tool for guiding further high-fidelity simulation or experimental validation.

In conclusion, the Mixture of Experts approach represents a promising and powerful direction for the future of scientific surrogate modeling, offering a clear path to harness the collective intelligence of a diverse and growing set of specialized AI tools.

\section*{acknowledgement}
We thank Rishikesh Ranade and Alexey Kamenev for preparing the expert prediction data for the DoMINO and FigConvNet models.

\bibliographystyle{unsrt}
\bibliography{ref}

\end{document}